\documentclass[conference]{IEEEtran}
\IEEEoverridecommandlockouts

\usepackage{cite}
\usepackage{amsmath,amssymb,amsfonts}
\usepackage{algorithmic}
\usepackage{graphicx}
\usepackage{textcomp}
\usepackage{xcolor}
\usepackage{multicol}
\usepackage{xcolor}
\usepackage{wrapfig}
\usepackage{hyperref}
\renewcommand{\arraystretch}{1.25}
\usepackage{xspace}
\usepackage{multirow}
\usepackage{graphicx}
\usepackage{booktabs}
\usepackage{amsmath}
\usepackage{amssymb}
\usepackage{makecell}
\usepackage[table]{xcolor}
\usepackage{colortbl}
\usepackage{subcaption}
\usepackage{circledsteps} % For text inside solid circle

\usepackage{listings} % For algos
\usepackage{xcolor}
\usepackage{tcolorbox}
\tcbuselibrary{breakable, listings, skins}

\newtcblisting{vlmprompt}{
  breakable,
  listing only,
  colback=gray!5,
  colframe=gray!50,
  boxrule=0.5pt,
  arc=1mm,
  left=2mm,
  right=2mm,
  top=2mm,
  bottom=2mm,
  listing options={
    basicstyle=\small\ttfamily,
    breaklines=true,
    columns=fullflexible,
    keepspaces=true,
    showstringspaces=false,
  }
}
\lstdefinestyle{pythonstyle}{
    language=Python,
    basicstyle=\footnotesize\ttfamily,
    keywordstyle=\color{blue},
    stringstyle=\color{red!60!black},
    commentstyle=\color{green!50!black},
    numbers=left,
    numberstyle=\tiny,
    stepnumber=1,
    numbersep=5pt,
    frame=single,
    breaklines=true,
    breakatwhitespace=true,
    showstringspaces=false,
    tabsize=2,
    captionpos=b
}

\definecolor{citecolor}{HTML}{D73F09} 

\hypersetup{
  colorlinks=true,
  citecolor=citecolor,
  linkcolor=citecolor,
  urlcolor=citecolor
}

\definecolor{PLWMcolor}{HTML}{0D5257}
\definecolor{GTRolloutColor}{RGB}{213, 51, 214}
\definecolor{PredRolloutColor}{RGB}{13, 82, 87}

\definecolor{best}{HTML}{7cbbbd}
\newcommand{\bestunderline}[1]{\textcolor{best}{\underline{\textcolor{black}{#1}}}}
\definecolor{secondbest}{HTML}{bfddde}

\author{Skand Peri, Hung Nguyen, Chanho Kim, Li Fuxin, Stefan Lee\\
Oregon State University}

\newcommand{\PLWMcolor}[1]{\textcolor{PLWMcolor}{#1}}

\newcommand{\mycircplwm}[1]{\Circled[fill color=PLWMcolor, inner color=white, outer color=white]{#1}}

\newcommand{\ours}{\PLWMcolor{\textbf{\textsc{3dPwm}}}}
\newcommand{\pformerplus}{ParticleFormer$^+$}

\newcommand{\xhdr}[1]{\vspace{0.5em}\noindent\textbf{#1}\quad}

\begin{document}

\title{\PLWMcolor{\textsc{3d Point World Models}}: Point Completion Enables More Accurate Dynamics Learning}

\maketitle

\begin{abstract}
Learning predictive models of the world enables robotic control through planning, potentially allowing robots to improvise solutions on new tasks. However, large video-based dynamics models lack explicit 3D spatial structure and suffer from geometrically inconsistent long-term rollouts with compounding errors. Emerging 3D dynamics models based on partial point clouds improve geometric consistency but remain sensitive to occlusions and accumulated prediction drift.
To address these challenges, we present 3D Point World Models (\ours{}) -- a task-agnostic world model that operates entirely in 3D space by first completing partial point clouds and then learning action-conditioned dynamics in this completed 3D scene. By operating on completed geometry, \ours{} enables reliable long-horizon rollouts and more accurate cost evaluation for model-based planning while supporting adaptation to new tasks. Experiments across different robotic embodiments and tabletop manipulation benchmarks demonstrate that \ours{} achieves substantially higher rollout accuracy compared with state-of-the-art 3D world models, including PointWorld, supports both open-loop and closed-loop planning, and enables successful sim-to-real transfer. \texttt{Webpage: \url{https://pvskand.github.io/3dpwm}}
\end{abstract}


\section{Introduction}
Embodied agents acting in physical environments benefit from having a 3D understanding of the world. To endow such agents with capabilities to reason about their surroundings and plan their actions accordingly, they should be able to not just interpret the current state of the world, but also predict the outcomes of an action that they would take. In other words, agents should learn a reliably accurate world model \cite{ha2018world} that facilitates planning in the real world.

Prior work in world model-based robot control has predominantly learned \textit{task-specific} world models from 2D images \cite{hafner2019dreamer, hafner2020dreamerv2, hafner2023dreamerv3, wu2023daydreamer, hansen2024tdmpc2}, using them for reinforcement learning or model based planning \cite{pets, nagabandi2020deep, hafner2019planet}. These models typically train both transition and reward functions, which must be fine-tuned or re-trained when adapting to new tasks. More recently, large video models \cite{agarwal2025cosmos, wu2024ivideogpt, ho2022video} have emerged by training on vast internet-scale video datasets to learn transition dynamics. However, these models face two critical challenges for real-world deployment. First, they are prone to hallucinating unrealistic or implausible future frames. Second, to enable action execution in physical environments, they require a learned inverse kinematics (IK) model to infer actions from generated videos \cite{liang2025video, du2023learning, zhou2024robodreamer}. Learning such robust and generalizable IK models remains challenging due to the inherent ambiguity in action-state correspondences -- for example, the same visual motion of a robot arm lifting an object could correspond to very different torque commands depending on the object’s (unobserved) mass or friction properties.

Another class of models that has been explored is 2.5D video models \cite{zhen2025tesseract, shang2025roboscape, ren2025gen3c} that jointly predict geometric properties such as depth and surface normals alongside RGB frames. They are more grounded in the physical 3D world, however, since such models process these additional features in a 2D fashion, they rely on implicit geometric reasoning and still remain prone to multi-view inconsistencies. For example, normals and depth inferred from one camera viewpoint may not match those inferred from another. Furthermore, owing to the dynamics model operating on partial RGB(-D) observations, both classes of models suffer from compounding rollout errors, which hamper reliable long-horizon prediction and planning.

\begin{figure}[t]
\centering
\includegraphics[width=\linewidth]{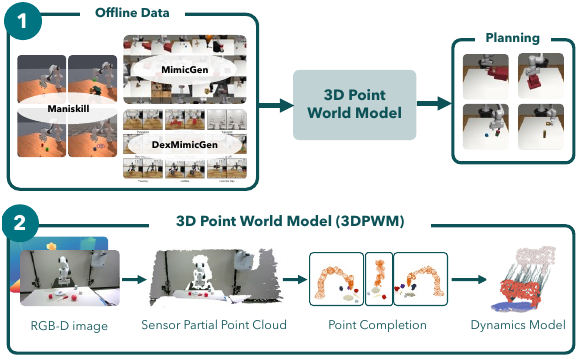}
\caption{\textit{Overview.} We propose \ours{}, a task-agnostic world model trained from demonstrations and deployed for planning via model-predictive control. Given a single-view RGB-D observation, the system constructs a partial point cloud, performs point completion, and then rolls out action-conditioned trajectories in 3D space.}
\label{fig:overview}
\vspace{-11pt}
\end{figure}

% Move on to motivating 3D dynamics models.
Recent work learns explicit 3D world models from partial point clouds \cite{huang2025particleformer, PCWM}, particles
\cite{ai2025review, allen2023graph, kim2024object}, or Gaussian splats \cite{zhang2024dynamic}, enabling planning by simulating how scene geometry evolves under actions. However, these approaches suffer compounding errors in occluded scenes due to partial-view inputs \cite{huang2025particleformer, PCWM,
zhang2024dynamic}. Meanwhile models trained on complete point clouds in simulation \cite{allen2023graph, li2018learning, kim2024object} struggle in real-world deployment where complete geometry is unavailable. We bridge this gap by designing a framework that integrates point cloud completion into 3D world modeling, enabling a world model to operate on complete 3D geometry in both simulation and the real world while retaining the benefits of training on complete geometry.
% To address these limitations, recent work has explored learning explicit 3D world models from partial point clouds \cite{huang2025particleformer, PCWM}, particles \cite{ai2025review, allen2023graph, kim2024object}, or Gaussian splats \cite{zhang2024dynamic}. Such models enable planning in 3D by simulating how the scene geometry evolves in response to actions. While these approaches produce more geometrically consistent rollouts, they remain susceptible to compounding errors in heavily occluded scenes due to their reliance on partial-view inputs \cite{huang2025particleformer, PCWM, zhang2024dynamic}. 

% On the other hand, models trained with complete point clouds in simulation \cite{allen2023graph, li2018learning,kim2024object} face challenges in real-world deployment because complete point clouds are difficult to obtain in real-world settings. In this work, we integrate point cloud completion into the world modeling pipeline to obtain complete geometry from partial-view inputs during real-world deployment.

% Proposed system

We propose \PLWMcolor{3D Point World Models} (\ours{}) -- a system to train 3D point cloud world models in a \textit{task-agnostic} fashion, allowing training on diverse (non-expert) trajectories. Different from prior work that learns from a vector latent space \cite{PCWM} or only from partial point clouds \cite{huang2025particleformer}, we emphasize the importance of complete 3D geometry and demonstrate that it is much easier for the dynamics model to reason about the center of gravity and impact in occluded areas. To that end, we develop a point completion pipeline tailored to robot manipulation settings and introduce an action-conditioned 3D world model that operates on complete 3D geometry, achieving state-of-the-art rollout accuracy across multiple robot embodiments Our key insight in this work is that complete point clouds offer better training regime for point clouds based world models, which in turn (a) facilitates reliable computation of cost functions during deployment for planning, and (b) adaptation to novel unseen tasks that were not present in the dynamics training distribution. Through results across various tasks, we show that the progress in scene completion \cite{chen2025sam3d} and, in particular, 3D point cloud completion \cite{Zhou2022SeedFormerPS, khademi2025point}, has reached sufficient maturity to be integrated into robotic systems. A high level overview of the system is shown in Figure \ref{fig:overview}.

\textbf{Contributions.} We summarize our main contributions: \newline
-- We introduce  \ours{}, a framework that integrates point completion with 3D world modeling to enable dynamics prediction over complete geometry, demonstrating state-of-the-art rollout performance across robot manipulation scenes. \newline
-- We provide a training recipe combining data augmentation, multi-step rollout losses and register tokens, that greatly improve the performance of \ours{}. %over prior work.
\newline
-- We show that \ours{}’s accurate long-horizon rollouts across different robot embodiments enable better closed-loop and open-loop planning than baselines, as well as adapting to a novel combination of tasks. \newline
-- Finally, we show effective sim-to-real transfer of the system on tabletop manipulation tasks where \ours{} performs $2.5\times$ better than baselines.

\section{Related Works}
\noindent \textbf{Point Cloud World Models.} Several prior works have explored learning dynamics models directly from meshes \cite{allen2023graph, fortunato2022multiscale} and point clouds using graph neural network (GNN), for both deformable and rigid-body systems \cite{kim2024object,shi2022robocraft, shi2023robocook, li2018learning, longhini2025cloth, tian2025diffusion}. More recently, to improve the accuracy of point cloud-based dynamics modeling, transformer architectures \cite{vaswani2017attention} have been adopted \cite{whitney2024modeling, huang2025particleformer}. In particular, ParticleFormer \cite{huang2025particleformer} introduced a transformer-based, object-centric point cloud dynamics model. However, these approaches differ from the proposed system in two key aspects: (a) the control capabilities demonstrated in prior methods are largely limited to visual planning on push-like tasks. In contrast, our approach supports more complex manipulation behaviors, including pick-and-place. (b) Prior work that operates on complete point clouds derived from meshes has primarily been evaluated in simulated environments \cite{li2018learning, fortunato2022multiscale, allen2023graph, kim2024object}. When training such dynamics models in real-world settings with partial point clouds \cite{shi2022robocraft, shi2023robocook, huang2025particleformer}, Chamfer distance is often used to handle points that disappear or appear due to occlusion and disocclusion. This results in weaker supervision signals for world model learning. In contrast, we train our world model using complete point clouds in simulation only and demonstrate sim-to-real transfer, when paired with a point cloud completion module, enabling effective 3D world modeling with complete geometry in real-world environments.

In PointWorld \cite{huang2026pointworld}, the authors train a single, large 3D point cloud dynamics model on real-world DROID dataset \cite{khazatsky2024droid} and show the effectiveness of their world model on several real-world tasks. Our model has a couple of notable differences with it: a) PointWorld operates on partial point clouds, resulting in rollouts with missing geometry when observations are heavily occluded, whereas our model reconstructs complete object geometry as long as objects are partially observable. We demonstrate that maintaining complete geometry during rollouts leads to better training for world models and enables more accurate cost functions for planning.
b) PointWorld demonstrates relatively shorter-horizon rollouts (30 steps for planning, 10 steps for rollout visualization), while our work demonstrates substantially longer-horizon rollouts (up to 300 steps) that remain stable and accurate. c) The manipulation scenarios in PointWorld primarily involve collision-free settings (except for the microwave door open task) in which an object is already grasped by the robot. In contrast, our work focuses on prehensile manipulation and supports planning from initial states before the robot has made contact with any object, in both open-loop and closed-loop control settings. 

\noindent Concurrent with \ours{} and PointWorld, MolmoMotion \cite{molmomotion} learns language-conditioned 3D object trajectories and demonstrates transfer to post-grasp pick-and-place policies. We instead learn robot-action-conditioned dynamics over completed 3D scenes, enabling long-horizon rollouts for explicit model-based planning.

\noindent \textbf{Latent World Models.}  While image based latent world models such as Dreamer\cite{hafner2019dreamer, hafner2019planet, hafner2020dreamerv2, hafner2023dreamerv3} and TDMPC \cite{Hansen2022tdmpc, hansen2024tdmpc2} learn task specific dynamics models, DINO-WM \cite{dinowm}, V-JEPA \cite{vjepa2} learns dynamics over frozen DINOv2 patch features and plans by optimizing action sequences toward goal features, sharing our offline, task-agnostic, reward-free setting. However, because its rollouts live in a 2D latent space, cost functions cannot be written over scene geometry, which is what enables \ours{}'s completion-based costs.

\noindent  \textbf{Point Cloud Completion.} Early work on point cloud completion focused on reconstructing a single object from a partial point cloud in a canonical object space \cite{yuan2018pcn, Huang_2020_CVPR,Zhou2022SeedFormerPS,khademi2023diverse}, which largely limited these methods to synthetic settings and made real-world deployment challenging, since real-world point clouds often contain multiple objects and are not aligned with a canonical object space. More recent work has shifted toward instance-level point cloud completion from partial point clouds of multi-object scenes, rather than single-object reconstruction \cite{Nie_2021_CVPR, tang2022point, Li_2023_ICCV, khademi2025point}. To the best of our knowledge, this is the first work to combine point cloud completion with 3D world modeling for real-world manipulation, enabling more accurate world models based on complete geometry.

\section{Methodology}
\label{sec:methodology}
\begin{figure*}[t]
\centering
\includegraphics[width=\textwidth]{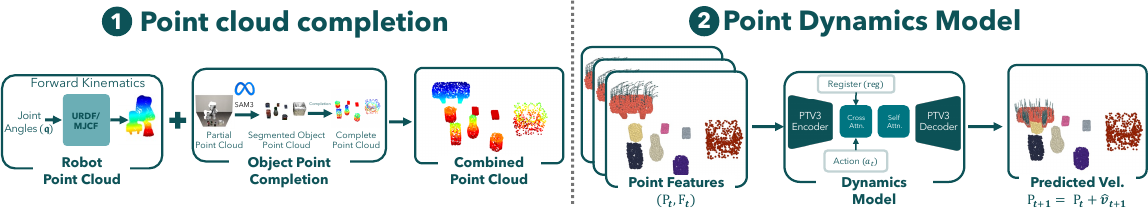}
\caption{\textit{System Overview.} \ours{} consists of \mycircplwm{1} \textit{Point completion}, which takes a single-view calibrated RGB-D input and robot joint angles to (a) sample end-effector points via forward kinematics on the robot URDF, (b) segment scene objects with SAM3 and complete their geometries using a point completion model \cite{khademi2025point}, and (c) merge robot and object points into a full scene point cloud; and \mycircplwm{2} \textit{Point dynamics}, which trains an action-conditioned model to predict next-step point velocities $\hat{v}$.}
\label{fig:system}
\vspace{-12pt}
\end{figure*}

\xhdr{\textbf{Problem Formulation.}}Our system takes in robot proprioception (with joint angles $\mathbf{q}$) and a single-view partial point cloud (obtained from an RGB-D camera) as the observation $\mathcal{O}$ at every step. Given this partial observation, we first complete the point cloud to obtain the complete observation of the visible scene $\mathcal{O}^{\text{complete}}$. Then, given an action $a \in \mathcal{A}$, we train the transition function $\mathcal{T}:\mathcal{O}^{\text{complete}} \times \mathcal{A} \rightarrow \mathcal{O}^{\text{complete}}$ that predicts the next observation. The goal is to use this learned dynamics model with a planner to roll out trajectories and optimize a desired cost objective.

\xhdr{\textbf{Obtaining Complete Point Clouds.}} While complete point clouds are available from object meshes during training in simulation, at deployment time we obtain complete point clouds as follows: 
% To obtain complete object point clouds, we first segment each visible object in the scene using SAM3 \cite{carion2025sam3segmentconcepts} with point prompts generated by VLM \cite{team2025gemini}, and back-project the depth map to recover partial object point clouds. We then apply point cloud completion \cite{khademi2025point} to produce completed object point clouds, ${\mathbf{P}_t^{\text{obj}}}$. The final observation is a combination of the robot end-effector points and complete object point clouds; $\mathbf{P}_t = \mathbf{P}_t^{\text{robot}} \cup \bigcup_{\text{obj}} \mathbf{P}_t^{\text{obj}}$.
We first segment each visible object in the scene using SAM3 \cite{carion2025sam3segmentconcepts} with point prompts generated by VLM \cite{team2025gemini}, and back-project the depth map to recover partial object point clouds. We then apply point cloud completion to produce completed object point clouds, ${\mathbf{P}_t^{\text{obj}}}$. We fine-tune the model from \cite{khademi2025point}, which is pre-trained on the ShapeNet dataset, using objects from the YCB \cite{YCB} dataset. This model improves upon previous point completion models in that it does not require knowledge of the complete object center, a shortcoming that prevents previous completion models from being used here. 

We revisit the data preprocessing and augmentation strategy. In~\cite{khademi2025point}, point cloud completion is formulated for upright objects in the scene by aligning each partial object point cloud input (extracted via instance segmentation) to a canonical upright orientation along the $y$-axis. While this simplifies learning, it restricts the ability of the model to generalize to dynamic scenarios in which objects may appear in arbitrary orientations, leading to degraded completion performance. To address this limitation, we apply rotation augmentation using all three axes ($x$, $y$, and $z$) during both pre-training on ShapeNet and fine-tuning on YCB~\cite{YCB}. Additionally, we apply random jittering to simulate depth noise from depth sensors. To further mitigate sensor depth noise, particularly near object boundaries and on glossy surfaces, we follow~\cite{zhang2025tapip3d} to clean up boundary-related depth noise and apply a median filter to suppress noisy depth values within each object.

To obtain complete robot point clouds, given $\mathbf{q}_t$ and the robot URDF, we use forward kinematics to compute the robot's mesh configuration and sample points from it to obtain the point cloud of the robot $\mathbf{P}_t^{\text{robot}} = \mathrm{FK}(\mathbf{q}_t, \mathrm{URDF})$. Since the action space operates in the end-effector frame, we only sample points from the end-effector mesh as shown in Fig. \ref{fig:system}. The final observation is a combination of the robot end-effector points and complete object point clouds; $\mathbf{P}_t = \mathbf{P}_t^{\text{robot}} \cup \bigcup_{\text{obj}} \mathbf{P}_t^{\text{obj}}$.

\xhdr{\textbf{3D Point World Model.}} Given the complete point cloud of the scene $\mathbf{P}_t$, we featurize each point with its current spatial location as well as both current and previous timestep velocities -- 
$\mathbf{F}_t = [\mathbf{x_t, y_t, z_t, v_t, v_{t-1}}].$ Since we train on datasets collected in simulation, we can obtain point-to-point correspondences across time to compute the velocities. In the real-world, these correspondences can be obtained via point tracking \cite{zhang2025tapip3d}. To train the dynamics model, this observation representation: ($\mathbf{P}_t, \mathbf{F}_t$) is encoded using a point cloud backbone. Specifically, we use Point Transformer V3 (PTV3) \cite{ptv3}, a state-of-the-art point cloud encoding architecture to obtain a point-based latent representation $\mathbf{z}_t \in \mathbb{R}^{M \times d}$, where $M$ is the number of points after downsampling and $d$ is the dimensionality of each point feature. At this bottleneck level, we induce action conditioning into the network by first encoding the action $a_t$ with a small MLP; $\mathbf{z}^\text{act}_t = \text{MLP}(a_t)$. We then perform cross-attention \cite{vaswani2017attention} where along with the action embedding, we also include a learnable \textit{register} token \cite{darcet2023visionregister}. These serve as keys and values, and the point features $\mathbf{z}_t$ serve as queries. This is to ensure that action-relevant points (i.e., corresponding to the robot) can attend to action commands, while action-irrelevant points (i.e., scene points not interacting with the robot) can attend to the register token ($\mathbf{reg}$), which effectively acts as a dummy key. We find this cross attention design with a register token to be important for performance. The cross attention layer is followed by self-attention layers to propagate the effect of the action across all scene points: 
\[
\tilde{\mathbf{z}}_t
= \mathrm{CrossAttn}\!\left(
\left[\mathbf{z}^\text{act}_t, \mathbf{reg}\right],\;
\mathbf{z}_t
\right), \qquad
\mathbf{z}'_t
= \mathrm{SelfAttn}(\tilde{\mathbf{z}}_t)
\]
Finally, the resulting point cloud representation is decoded using PTV3’s point decoder to obtain a feature representation at the original point cloud resolution. A lightweight MLP is then applied to the per-point features to predict per-point velocities at the next timestep, $\hat{v}_{t+1}.$
The next timestep point cloud can be obtained by:
\begin{equation}
    \label{eqn:delta_prediction}
    \mathbf{\hat{P}}_{t+1} = \mathbf{P}_{t} + \hat{\mathbf{v}}_{t+1}
\end{equation}
Long-horizon rollouts are produced by recursively applying the dynamics model and Eq.\ref{eqn:delta_prediction} where the input observation for timestep $t+i$ is:
\begin{equation}
    (\mathbf{\hat{P}}_{t+i}, \mathbf{\hat{F}}_{t+i}) = (\mathbf{\hat{P}}_{t+i}, [\hat{\mathbf{x}}_{t+i}, \hat{\mathbf{y}}_{t+i}, \hat{\mathbf{z}}_{t+i}, \hat{\mathbf{v}}_{t+i}, \hat{\mathbf{v}}_{t+i-1}])
\end{equation}

\xhdr{\textbf{Training Objective.}} Given access to point-to-point correspondences in our dataset, we train the \ours{} using a Huber loss \cite{huber1992robust} between the predicted and ground-truth per-point velocities. This provides direct supervision on the dynamics, in contrast to prior work \cite{huang2025particleformer} that relies on Chamfer or Hausdorff distances. We empirically validate the benefit of this formulation in Section \ref{sec:expts}. Rather than providing single step supervision, we train \ours{} based on multi-step rollouts -- iteratively applying our model to produce the following $H$ time steps induced by an action sequence $a_t, ..., a_{t+H}$:
\begin{equation}
\mathcal{L} = \frac{1}{H} \sum_{h=1}^{H} \text{Huber}\Big(\hat{\mathbf{v}}_{t+h}, \mathbf{v}_{t+h}\Big),
\end{equation}
We find that the multi-step rollout loss functions significantly reduce compounding errors (Sec.~\ref{sec:ablation}). 

\xhdr{\textbf{Planning with \ours{}.}} We employ \ours{} in a model predictive control framework (MPC) using a sampling based MPPI planner \cite{williams2017model}. We roll out $\mathcal{K}$ trajectories at each step and optimize with a pre-defined cost function. This cost function is defined in the observation space of the complete point cloud $\mathbf{P}_t$, similar to how a reward function for a downstream task is written with a simulator.

\section{Experiments}
\label{sec:expts}
We show the efficacy of \ours{} along 4 fronts: (1) Can \ours{} lead to reliable and accurate long-horizon rollouts on diverse embodiments?; (2) How well can \ours{} be used to effectively plan for robotic manipulation tasks?; (3) Can \ours{} be used for adaptation to unseen tasks; and (4) Can \ours{} be transferred to real-world manipulation tasks?

\textbf{Data Collection.} For training the dynamics modeling, we collect data from a mix of expert demonstrations and augment the demos with random actions to cover a larger portion of the state space. Specifically, with a probability of 0.2, we induce randomness in the expert trajectory by uniformly selecting a timestep $i$ %\in \texttt{0, len(traj)}$ 
and executing random action from step $i$. All actions are recorded in delta end-effector space. = Instead of completely using random trajectories from the beginning, our rationale to have this kind of augmentation is to induce novel collision dynamics that are unseen in the expert demonstration -- such as trying to execute a task and failing mid-way. Early on in the project, we observed that this kind of augmentation led to a dynamics model that was more performant on planning. Specifically on \texttt{PickYCB}, this data augmentation boosted the performance of closed-loop planning from 58\% to 65\%. On \texttt{PickCube} the difference was similar ($\sim$ 83\% $\rightarrow$ 90\%). Additional details are in Sec. \ref{appdx:data_generation}.

%We collect point cloud data by replaying demonstration trajectories in simulation, obtaining complete point clouds by sampling from object and robot meshes. All actions are recorded in delta end-effector space. Since expert demos exhibit limited state-space diversity, we inject randomness by randomly select a timestep and thereafter execute uniformly sampled actions: for 20\% of trajectories. This yields partially successful trajectories that introduce diverse interaction and collision dynamics valuable for training. Additional details are in Sec. \ref{appdx:model_hyperparameters}.

\begin{figure*}[t]
\centering
\includegraphics[width=0.95\textwidth]{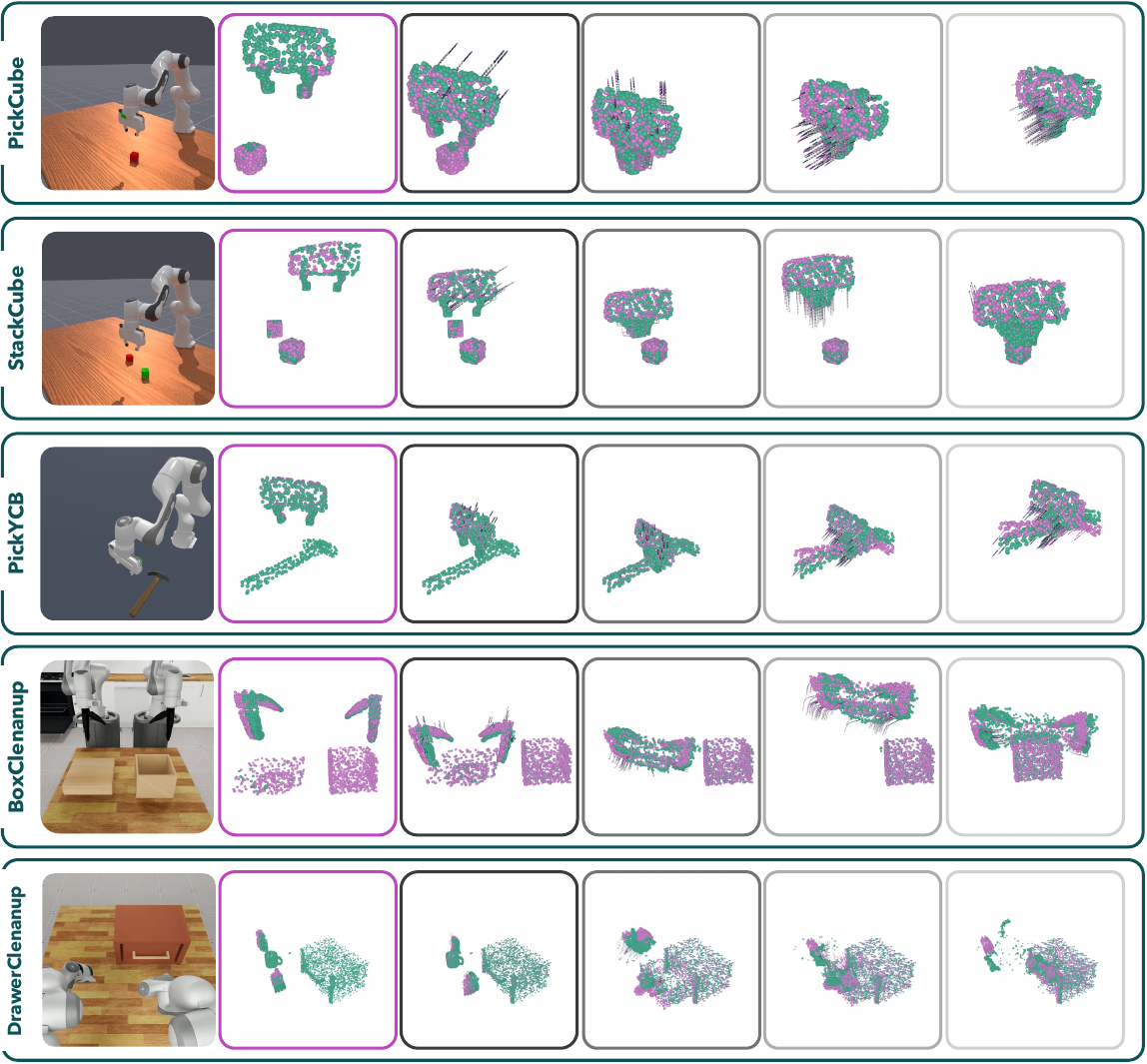}
\caption{\textit{Qualitative rollout results of \ours{}.} Action-conditioned rollouts of \ours{} over time for five tasks. Ground-truth trajectories are shown in \textcolor{GTRolloutColor}{\textbf{purple}}, while predicted rollouts are shown in \textcolor{PredRolloutColor}{\textbf{teal}}. The second column (with \textcolor{GTRolloutColor}{purple} border) shows the initial observed point cloud, and all subsequent point clouds are generated by rolling out the model using actions from the ground-truth trajectory. A higher overlap between the \textcolor{GTRolloutColor}{purple} and \textcolor{PredRolloutColor}{teal} points indicate better accuracy of world model rollouts.}
\label{fig:qualitative-rollout}
\vspace{-10pt}
\end{figure*}

% To show the efficacy of \ours{}, we answer the following questions: (1) Can \ours{} lead to reliable and accurate long-horizon rollouts on diverse embodiments?; (2) How well can \ours{} be used to effectively plan for robotic manipulation tasks?; (3) Can \ours{} be used for adaptation to unseen tasks; and (4) Can \ours{} be transferred to real-world manipulation tasks?

\vspace{-2pt}
\subsection{\textbf{Evaluating accuracy of long-horizon rollouts.}} 
\label{sec:rollout-accuracy}
\vspace{-2pt}

\textbf{Setting.} To evaluate \ours{}’s rollout capability, we train on seven datasets spanning two end-effector embodiments (Franka parallel gripper and GT-1 dexterous hand). For Franka, we use \texttt{PickCube}, \texttt{StackCube}, and \texttt{PickYCB} from ManiSkill \cite{Tao2024ManiSkill3GP}, and \texttt{MugCleanup} from MimicGen \cite{mandlekar2023mimicgen}. The PickYCB task requires the robot to be able to pick up any of the 78 objects from YCB dataset, and is considered a difficult task for planning. For the dexterous suite, we use \texttt{BoxCleanup}, and \texttt{DrawerCleanup} from DexMimicGen \cite{jiang2025dexmimicgen}. Additional details are described in Appdx. \ref{appdx:environments}. \newline 
\textbf{Baselines.} We compare \ours{} against three baselines: \newline
-- \textit{PointWorld}: PointWorld \cite{huang2026pointworld} is the closest work to ours, but differs in two key aspects: 1) it operates on partial point clouds and 2) it uses action flows rather than end-effector (EE) actions as its action space. For a fair comparison, we convert our EE actions into action flows using inverse kinematics. \newline
-- \textit{\pformerplus}: A modified version of ParticleFormer \cite{huang2025particleformer} that adopts the same PTV3-based encoder–decoder as ours for a fair comparison. Note that this is a significantly larger network than the one in \cite{huang2025particleformer} which only had 3 self-attention layers. The model is trained using the hybrid loss proposed by the authors and operates on object-centric partial point clouds. \newline
-- \textit{Partial \ours{}}: A variant of \ours{} in which the point completion module is removed; equivalently, this corresponds to \pformerplus without object-centric features and trained with Chamfer and Hausdorff losses as done in \cite{huang2025particleformer}. \newline
For all methods, the model observes the first three point clouds of each test trajectory and then rolls out to predict future point clouds, conditioned on the actions from the test demonstration.
\newline
\textbf{Metrics.} We report the $\mathcal{L}_1$ Chamfer distance between rolled-out and ground-truth point clouds in Tab. \ref{tab:rollout_metrics}. However, this global metric does not directly reflect planning utility, as drift in critical points (e.g., the tool center point or an object's center of mass) can be masked by averaging.
We therefore additionally evaluate downstream task rewards on rolled-out point clouds, comparing them to ground-truth simulator rewards using Pearson correlation ($r$). For ManiSkill tasks, we use the benchmark's shaped rewards; for MimicGen and DexMimicGen, we extract reward functions using VLM (details in Appdx. \ref{appdx:cost_fn}). For all partial baselines, we use the visible-point mean as the CoM, providing a fair estimate under the available partial observation without assuming access to the privileged unobserved geometry. \newline
\textbf{Results.} Across all tasks, \ours{} achieves significantly lower Chamfer distance (average rollout lengths reported in Tab. \ref{tab:rollout_metrics}). \pformerplus, Partial \ours{}, and PointWorld are competitive with each other across different tasks, but all exhibit substantially higher rollout errors than \ours{}. PointWorld is competitive on the shorter parallel gripper tasks, but its performance degrades on long-horizon tasks and with the dexterous hand end-effector. Similarly, its reward correlation drops sharply on the dexterous hand tasks. Overall, these results show that learning dynamics from partial point clouds is challenging as increasing occlusions limit the model's ability to infer complete scene geometry; explicit completion removes the need for the dynamics model to implicitly reason about occluded points. Consistently, \ours{} outperforms all baselines on reward corr., further corroborating the benefit of point completion. Qualitative results are shown in Fig. \ref{fig:qualitative-rollout}. Results with additional baselines, including a GNN-based baseline, are provided in the Appendix.
\vspace{-2pt}
\subsection{\textbf{Model-based Planning.}} 
\label{sec:planning}
\begin{figure}[t]
%\vskip -0.1in
\centering
\includegraphics[width=\linewidth]{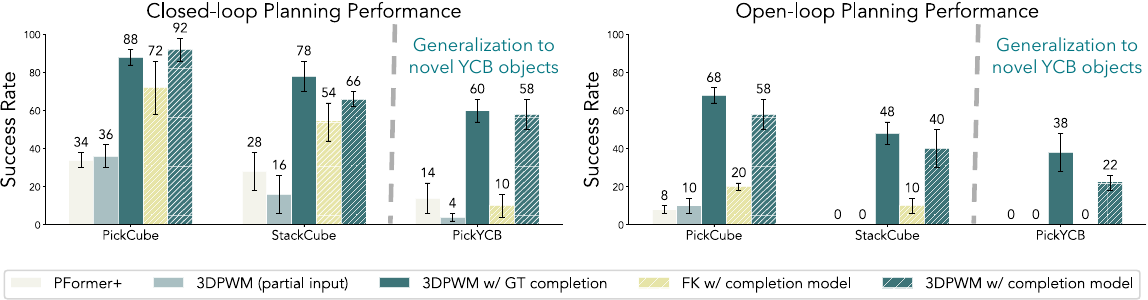}
\caption{\textit{Planning success rate}: \textbf{(Left)} Closed-loop planning success.  % on \texttt{PickCube}, \texttt{StackCube}, and \texttt{PickYCB}, 
%comparing \ours{} (with and without the completion module), Partial \ours{}, and \pformerplus.
\ours{} achieves the highest success across all tasks. \textbf{(Right)} Open-loop planning success on the same tasks, where \ours{} significantly outperforms prior methods due to higher-fidelity rollouts.}
\label{fig:planning}
\vspace{-15pt}
\end{figure}

\vspace{-2pt}
\textbf{Setting.} We evaluate \ours{} under model-based planning using MPPI \cite{williams2017model} in both \textit{open-loop} and \textit{closed-loop} settings on \texttt{PickCube}, \texttt{StackCube}, and \texttt{PickYCB}. A key motivation for learning a dynamics model is to use it as a proxy simulator in the real world, where ground-truth observations are unavailable, making accurate open-loop control important. In the open-loop setting, the system observes only the initial point cloud and plans actions for the entire horizon, serving as a stress test to assess whether \ours{} can function as a robust simulator for planning. To obtain the cost function for planning, following works like Eureka \cite{ma2023eureka}, and Dr. Eureka \cite{ma2024dreureka} - we prompt Gemini 2.5 Flash VLM \cite{comanici2025gemini} to generate a cost function given the annotated objects in the scene. \newline
\textbf{Baselines.} Beyond \pformerplus and Partial \ours{}, we evaluate \ours{} with ground-truth and fine-tuned point completion to quantify the impact of learned completion on planning success. This comparison quantifies the impact of learned point completion on planning success. In addition, we introduce an analytic transition model (denoted by \texttt{FK}) that uses forward kinematics to predict future EE poses and compute their costs. This comparison evaluates the importance of learning a dynamics model for the task.  \newline
\textbf{Metrics.} We report success rates over 100 MPPI runs, rolling out 200 samples with horizon $H=10$ at each step, for a maximum of 100 time steps. \newline
\textbf{Results.} Fig. \ref{fig:planning} reports performance on three tasks under closed-loop and open-loop planning.
\newline
\underline{\textit{Closed-loop planning.}} With GT completion, \ours{} achieves 88\% success on \texttt{PickCube}, decreasing to ~78\% on \texttt{StackCube} and ~60\% on \texttt{PickYCB} as task complexity increases. Learned point completion introduces a 2-10\% drop on the harder tasks due to noisy completions (see Appendix \ref{appdx:failure_cases}). Both \pformerplus and Partial \ours{} exhibit low success rates on partial point clouds, with \pformerplus performing slightly better, likely due to its explicit object-centric inductive bias. Interestingly, the analytical forward kinematics model combined with point completion achieves good success rates on both \texttt{PickCube} and \texttt{StackCube}. However, its performance decreases on the more diverse and challenging \texttt{PickYCB} task. This suggests that while simple kinematic models can be sufficient for relatively simple objects and interactions, handling a diverse set of objects benefits from reasoning about the consequences of multiple MPPI-generated actions over longer horizons, where \ours{} provides a clear advantage. \newline
\underline{\textit{Open-loop planning.}} All methods show a substantial drop relative to closed-loop, but \ours{} with ground-truth completion—and to a lesser extent with learned completion—still achieves non-trivial success rates. In contrast, \pformerplus and Partial \ours{} largely fail (1/20 on \texttt{PickCube}, 0\% elsewhere), due to low-quality rollout predictions leading to inaccurate cost estimation.

\begin{table}[t]
\begin{center}
\caption{Chamfer distance and reward correlation across embodiments and tasks (\textcolor{best}{Best} and \textcolor{secondbest}{second best} methods are highlighted).}
\centering
\label{tab:rollout_metrics}
\renewcommand*{\arraystretch}{1.0}
\setlength{\tabcolsep}{2.5pt}
\footnotesize
\resizebox{\linewidth}{!}{
\begin{tabular}{l l c 
>{\centering\arraybackslash}p{3.5em}
>{\centering\arraybackslash}p{3.5em}
>{\centering\arraybackslash}p{3.5em}
>{\centering\arraybackslash}p{3.5em}
|
>{\centering\arraybackslash}p{3.5em}
>{\centering\arraybackslash}p{3.5em}
>{\centering\arraybackslash}p{3.5em}
>{\centering\arraybackslash}p{3.5em}}
\toprule

\multirow{2}{*}[-2em]{\textbf{\makecell{End-effector}}} 
& \multirow{2}{*}[-2em]{\textbf{Task}} 
& \multirow{2}{*}[-2em]{\textbf{\makecell{Avg. \\ len.}}} 

& \multicolumn{4}{c}{\textbf{Chamfer L1 ($\downarrow$)}} 
& \multicolumn{4}{c}{\textbf{Reward Corr $(r)$ ($\uparrow$)}} \\

\cmidrule(lr){4-7} 
\cmidrule(lr){8-11}

& & 
& PF$^+$ 
& \makecell{Partial\\\ours{}} 
& \makecell{Point\\World} 
& \ours{} 
& PF$^+$ 
& \makecell{Partial\\\ours{}} 
& \makecell{Point\\World}  
& \ours{} \\

\midrule

\multirow{4}{*}{\makecell{Parallel\\Gripper}}
 
& PickCube & 66
& 0.080 
& 0.085 
& \cellcolor{secondbest}0.071
& \cellcolor{best}0.009
& 0.61 
& 0.57 
& \cellcolor{secondbest}0.69
& \cellcolor{best}0.86\\

& StackCube & 99
& \cellcolor{secondbest}0.090 
& 0.097 
& 0.094
& \cellcolor{best}0.014
& 0.55 
& 0.55 
& 0.53
& \cellcolor{best}0.83 \\

& Pick YCB & 143
& 0.089 
& 0.089 
& \cellcolor{secondbest}0.064
& \cellcolor{best}0.028
& 0.53 
& 0.51 
& \cellcolor{secondbest}0.68
& \cellcolor{best}0.85 \\

& Mug Cleanup & 336
& \cellcolor{secondbest}0.130 
& 0.132 
& 0.174
& \cellcolor{best}0.022
& 0.40 
& \cellcolor{secondbest}0.45 
& 0.35
& \cellcolor{best}0.79 \\

\midrule

\multirow{2}{*}{\makecell{Dexterous\\Hand}}

& Box Cleanup & 218
& \cellcolor{secondbest}0.132 
& 0.149 
& 0.211
& \cellcolor{best}0.022
& 0.30 
& \cellcolor{secondbest}0.32 
& 0.11
& \cellcolor{best}0.83 \\

& Drawer Cleanup & 274
& 0.273 
& \cellcolor{secondbest}0.246 
& 0.351
& \cellcolor{best}0.058
& \cellcolor{secondbest}0.29 
& 0.25 
& 0.04
& \cellcolor{best}0.77 \\

\bottomrule
\end{tabular}}
\end{center}
\vspace{-12pt}
\end{table}

\vspace{-2pt}
\subsection{\textbf{Adaptation to \textit{unseen} long-horizon tasks.}} 
\label{sec:adaptation}
\vspace{-2pt}
\begin{figure}[t]
\centering
\includegraphics[width=0.97\linewidth]{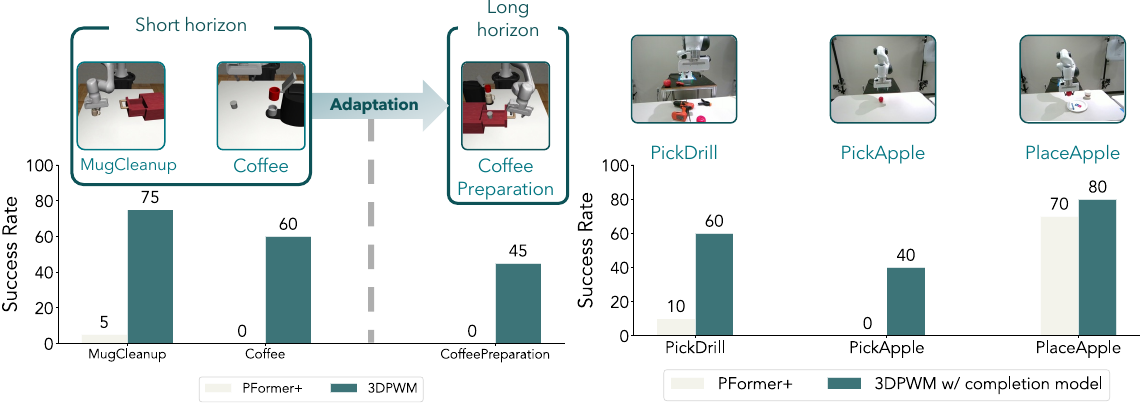}
\caption{\textbf{(Left)} \textit{Adaptation to unseen tasks}: We train \ours{} dynamics on demonstrations generated on \texttt{MugCleanup} and \texttt{Coffee} tasks and show that it can be used to adapt to a much long-horizon task of \texttt{CoffeePreparation} and achieving $45\%$ success rate. \textbf{(Right)} \textit{Sim2Real}: We evaluate sim-to-real transfer of \ours{} on \texttt{PickApple}, \texttt{PickDrill}, and \texttt{PlaceApple}. The static third-person camera setup and objects are shown in the top-right. While \ours{} and \pformerplus perform similarly on the easier placing task, \ours{} significantly outperforms on the picking tasks.}
\label{fig:sim2real}
\vspace{-12pt}
\end{figure}
\textbf{Setting.} We train \ours{} on MimicGen \cite{mandlekar2023mimicgen} demonstrations for the \texttt{Coffee} and \texttt{Mug Cleanup} tasks and evaluate on \texttt{Coffee Preparation}, a long-horizon task requiring the robot to place a mug on the coffee machine, retrieve a pod from a drawer, load it, and close the lid (Fig \ref{fig:sim2real}). This tests whether \ours{} can compose individually learned dynamics into extended task sequences. We use \pformerplus{} as a baseline to assess whether point completion aids generalization to new tasks. \newline
\textbf{Results.} On short-horizon tasks, \ours{} achieves 75\% on \texttt{MugCleanup} and 60\% on \texttt{Coffee}, with the latter's lower performance likely due to precise placement requirements. \pformerplus{} performs poorly on both due to incorrect dynamic predictions. On the long-horizon \texttt{CoffeePreparation}, chaining the subtasks leads to a 15\% drop for \ours{} from compounding execution errors, while \pformerplus{} fails entirely.
\vspace{-2pt}
\subsection{\textbf{Sim2Real transfer of \ours{}.}} 
\label{sec:sim2real}
\vspace{-2pt}
\xhdr{\textbf{Tasks.}} We evaluate sim-to-real transfer on tabletop \textit{Pick} and \textit{Place} tasks using a Franka Panda with a parallel-jaw gripper: \texttt{PickApple}, \texttt{PickDrill}, and \texttt{PlaceApple} (placing a toy apple onto a plate), with randomly placed distractors. Task setups and objects are shown in Fig. \ref{fig:sim2real}.\newline
\textbf{System Design.} We use a single-view RGB-D observation from an Intel RealSense D455 camera, extrinsically calibrated to align with the Franka base and ManiSkill2 coordinate system \cite{gu2023maniskill2}. Target objects are segmented using SAM3 \cite{carion2025sam3segmentconcepts} with text prompts and back-projected into 3D partial point clouds, which are then completed using a fine-tuned completion module \cite{khademi2025point}. For dynamics, we deploy the best \texttt{PickYCB} checkpoint from simulation \textit{without} real-world fine-tuning. Planning uses MPPI with 30--50 step open-loop rollouts executed in chunks. Additional details are in Appd. \ref{appdx:model_hyperparameters}; robustness to camera viewpoints and lighting is reported in Appdx. \ref{appdx:lighting_camera}.\newline
\textbf{Baselines \& Metrics.} Since Partial \ours{} performs comparably or worse than \pformerplus in simulation, we compare only against \pformerplus in real-world experiments, reporting success rates over 10 trials per \texttt{{method, task}} pair.\newline
\textbf{Results.} Results are shown in Figure \ref{fig:sim2real}. On \texttt{PlaceApple}, both methods achieve comparable success (70--80\%), as the plate's large support surface yields low placement failure rates. On \texttt{PickApple} and \texttt{PickDrill}, \ours{} substantially outperforms \pformerplus, which we attribute to improved object geometry from point completion. Remaining failures primarily arise from biased center-of-mass estimation (see Appdx.~\ref{appdx:failure_cases}).

\section{Ablation Study \& Discussion}
\label{sec:ablation}
% \vspace{-3pt}
We examine two key design choices in \ours{}: the role of \textit{register tokens} and the multi-step rollout length for training. \newline
\textbf{Register Tokens.} We ablate the number of register tokens \cite{darcet2023visionregister} ($\#\text{reg} \in \{0, 1, 2\}$) used as keys in cross-attention alongside the action encoding $\mathbf{z}^{\text{act}}_t$ (Table \ref{tab:ablation}). Increasing from $\#\text{reg}=0$ to $\#\text{reg}=1$ yields substantial gains, while additional tokens show diminishing returns. We use $\#\text{reg}=1$ for all experiments. As discussed in Sec. \ref{sec:methodology}, we hypothesize that register tokens help the model focus on task-relevant points in $\mathbf{z}_t$ during cross-attention. \newline
\textbf{Multi-step supervision.} We ablate the supervision horizon $H \in \{3,5,10,15\}$ (Table \ref{tab:ablation}). Longer horizons consistently improve dynamics modeling at the cost of increased training time due to autoregressive rollouts. We use $H=15$ for all experiments.

% \begin{table}[t]
% \centering

% \begin{minipage}{0.49\linewidth}
% \centering
% \caption{Effect of Register token \cite{darcet2023visionregister} in \ours{} on Chamfer dist. between GT and rollouts.}
% \label{tab:register_token}
% \renewcommand{\arraystretch}{1.15}
% \setlength{\tabcolsep}{2pt}
% \begin{tabular}{lccc}
% \toprule
% \textbf{Task} & \textbf{$\#$reg=0} & \textbf{$\#$reg=1}  & \textbf{$\#$reg=2} \\
% \midrule
% PickCube    & 0.013 & 0.009 & 0.009\\
% PickYCB     & 0.052 & 0.028 & 0.027 \\
% Mug Cleanup & 0.178 & 0.022 & 0.020 \\
% \bottomrule
% \end{tabular}
% \end{minipage}
% \hfill
% \begin{minipage}{0.49\linewidth}
% \centering
% \caption{Performance of \ours{} across horizon $H$ on Chamfer dist. between GT and rollouts.}
% \label{tab:multi_step_ablation}
% \renewcommand{\arraystretch}{1.15}
% \setlength{\tabcolsep}{2pt}
% \begin{tabular}{lcccc}
% \toprule
% \textbf{Task} & \textbf{$H=3$} & \textbf{$H=5$} & \textbf{$H=10$} & \textbf{$H=15$} \\
% \midrule
% PickCube   & 0.068 & 0.038 & 0.021 & 0.009 \\
% PickYCB    & 0.161 & 0.117 & 0.044 & 0.028 \\
% Mug Cleanup & 0.102 & 0.036 & 0.028 & 0.022 \\
% \bottomrule
% \end{tabular}
% \end{minipage}
% \vspace{-12pt}
% \end{table}

\begin{table}[t]
\centering
\caption{Ablation studies of \ours{} using Chamfer distance between GT and rollouts. We evaluate the effect of the \# register tokens and prediction horizon $H$. The best-performing configuration is \cellcolor[best]{shaded}, while the closest competitive configuration without a significant performance loss is \bestunderline{underlined}.}
\label{tab:ablation}
\renewcommand{\arraystretch}{1.1}
\setlength{\tabcolsep}{3pt}
\begin{tabular}{lccc|cccc}
\toprule
& \multicolumn{3}{c|}{\textbf{\# Register tokens}}
& \multicolumn{4}{c}{\textbf{Horizon $H$}} \\
\textbf{Task}
& \textbf{0} & \textbf{1} & \textbf{2}
& \textbf{3} & \textbf{5} & \textbf{10} & \textbf{15} \\
\midrule
PickCube
& 0.013 & \cellcolor{best}{0.009} & \cellcolor{best}{0.009}
& 0.068 & 0.038 & 0.021 & \cellcolor{best}{0.009} \\
PickYCB
& 0.052 & \bestunderline{0.028} & \cellcolor{best}{0.027}
& 0.161 & 0.117 & 0.044 & \cellcolor{best}{0.028} \\
Mug Cleanup
& 0.178 & \bestunderline{0.022} & \cellcolor{best}{0.020}
& 0.102 & 0.036 & 0.028 & \cellcolor{best}{0.022} \\
\bottomrule
\end{tabular}
\vskip -0.1in
\end{table}
\noindent \textbf{Computational time.} On a standalone RTX-4090, planning one action takes $\sim$1.07s, with tasks requiring 4-5 open-loop chunks of 20 actions. The main latency source is sequential dynamics rollout in observation space, which could be sped up via CUDA kernels supporting chunked multi-step prediction. Training takes 12--24 hours on a single L40 GPU, depending on dataset size, with larger-point datasets (\texttt{MugCleanup}) and higher-dimensional action spaces (\texttt{TwoArmBoxCleanup}, \texttt{TwoArmDrawerCleanup}) requiring closer to 24 hours.

\section{Limitations}
\label{sec:limitations}
% \vspace{-5pt}
% Our system, \ours{}, is composed of several modular components -- SAM3-based object segmentation, object-centric point completion, dynamics modeling, and planning via model rollouts. Failures in any stage can propagate through the pipeline, leading to cascading errors. \newline
% \textbf{SAM failures.} A common source of failure in our Sim2Real experiments was open-vocabulary object segmentation. SAM3 often failed to detect objects that were partially occluded or held by the gripper, accounting for 4/12 failures. \newline
% \textbf{Point Completion Failures.} Point completion was the most frequent failure mode (5/12 failed trials). The object partial point clouds obtained by back-projecting sensor depth were noisy and out of distribution for the completion module, which was trained on individual YCB objects \cite{YCB} without scene-level occlusions. Training with scene-centric partial inputs would likely improve robustness. \newline
% \textbf{Latency.} The modular design introduces latency at each stage. Despite the heavy segmentation backbone (848M params.), the main bottleneck is planning with the dynamics model due to sequential rollouts. We perform open-loop planning in chunks, which works well for pseudo-dynamic tasks but limits extension to fully dynamic manipulation \cite{islam2021provably}. This latency may be mitigated by chunked multi-step prediction \cite{huang2026pointworld} or lower-dimensional parametric action representations.
\ours{} combines several modular components—VLMs, foundation segmentation models (SAM3), point completion, dynamics modeling, and model-based planning. This modular design enables flexible integration of these components, but also leaves room to improve robustness and efficiency. In our Sim2Real experiments, segmentation occasionally struggled with partially occluded or gripper-held objects, and point completion would likely benefit from training on more diverse, scene-level partial point clouds. Additionally, the sequential dynamics rollouts used for planning introduce latency and currently favor quasi-static tasks. Addressing these limitations—for example, through scene-centric completion training, chunked multi-step prediction \cite{huang2026pointworld}, and lower-dimensional action representations—could further improve the robustness and scalability of \ours{}.

\section{Conclusion}
% \vspace{-5pt}
In this work, we demonstrate that world models based on explicit point cloud completion lead to improved rollout quality. \ours{} achieves better predicted geometry, reward correlation, and downstream planning performance across simulated and real tasks, including adaptation to new long-horizon tasks. Our analysis suggests tighter integration of perception components and reduced planning latency are critical directions for further improvement.

\section{Author Contributions}
\textbf{Skand:} Led the project. Implemented the data generation pipeline for point completion and dynamics model training, trained dynamics model, and wrote the planner. Primary paper writer including figures. 

\textbf{Hung Nguyen:} led the point completion training, and helped with sim-to-real experiments in integrating SAM3, with point completion module.

\textbf{Chanho Kim:} helped with writing of the paper, participated in discussions throughout the project and provided inputs on methodology.  

\textbf{Li Fuxin:} contributed to manuscript preparation and revision, provided guidance on methodology, and offered critical feedback and supervision throughout the project.

\textbf{Stefan Lee:} contributed to manuscript preparation and revision, provided guidance on methodology, and offered critical feedback and supervision throughout the project.

\section{Change Log}
\begin{itemize}
    \item Added \textbf{PointWorld} as a rollout baseline.
    \item Added an analytic \textbf{forward-kinematics (FK)} planning baseline.
    \item Revised related work to include newer 3D and 2D latent world models.
    \item Increased simulated planning evaluation from \textbf{20 to 100 runs}.
    \item Added \textbf{VLM-generated cost functions} for simulated planning experiments, replacing the previously hand-designed cost functions.
\end{itemize}

\bibliographystyle{IEEEtran}
\bibliography{references}

\clearpage 

\section{Appendix}
\subsection{Data Generation}
\label{appdx:data_generation}

\begin{figure}[!h]
\centering
\includegraphics[width=0.9\linewidth]{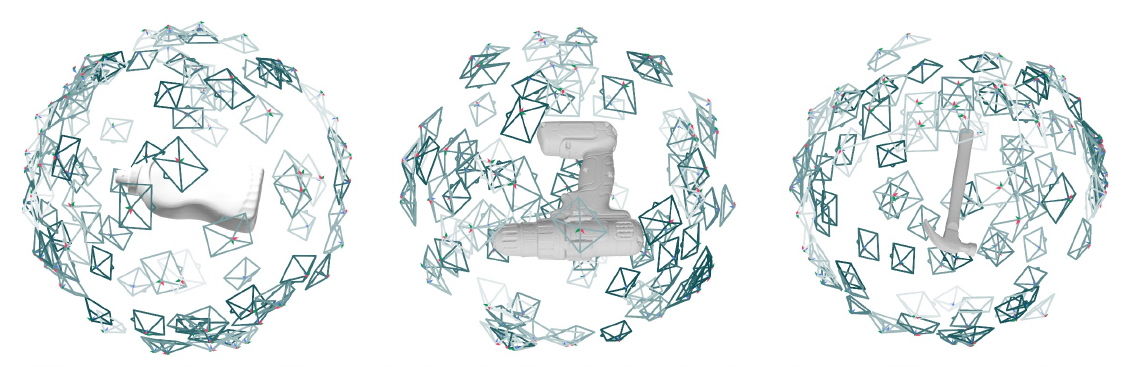}
\caption{\textit{Point completion Data Generation}: Given the mesh/point cloud of the object, we render depth maps from 100 random views around a sphere of 1 unit looking at $(0, 0,0)$, with the object in centered at the origin.}
\label{fig:point_completion_views}
\end{figure}

\textbf{Point completion.} We consider 100 random viewpoints for each object from the YCB dataset from which we obtain the partial point clouds inputs. A sample of the random views on three YCB objects is shown in Figure \ref{fig:point_completion_views}.
 
\textbf{Dynamics modeling.} For all tasks across ManiSkill and DexMimicGen, we write a custom wrapper to obtain the complete point clouds from the simulation itself by sampling points from objects/parts/geom meshes. Further, to handle point-to-point correspondences, we only sampled the points at the cannonical frame and applied rigid body pose transformations for all the desired entities in the scene. This significantly boosted the training data collection time as well as reduced the need to store point clouds at each step (only storing one cannonical point cloud and $\texttt{len(traj)}$ transformation matrices).
\subsection{Model Details \& Hyperparameters}
\label{appdx:model_hyperparameters}

\xhdr{\ours{} dynamics modeling.} For training the dynamics model, we adopt the PTV3 architecture \cite{ptv3}, replacing batch normalization with layer normalization following Sonata \cite{wu2025sonata}. To better suit the smaller dataset size, we reduce the depth of each of the four PTV3 encoding stages from the default configuration to \texttt{[2,2,2,2]}. Relative positional encoding (RPE) helps to capture spatial relationships between the points, but enabling it throughout the network makes it hard to use FlashAttention \cite{dao2022flashattention, dao2023flashattention2} to speed up the training process. Instead, we enable RPE in alternating encoder and decoder blocks. Accordingly, we adjust the patch size for neighborhood computation from 256 in the layers where RPE is disabled to 48 when RPE is enabled.
The bottleneck per-point feature dimension is set to $d = 256$, while the number of input points $M$ depends on the grid resolution and scene complexity, typically ranging from 200 to 400 points. For action conditioning, both the \texttt{CrossAttn} and \texttt{SelfAttn} modules consist of three layers with eight attention heads each. The decoder outputs 128-dimensional per-point features, which are passed through a four-layer MLP to predict per-point velocities. Wherever applicable, we employ FlashAttention \cite{dao2022flashattention, dao2023flashattention2} across all attention modules in PTV3, including the cross- and self-attention layers, to improve training efficiency. We provide detailed hyperparameters for the completion module, dynamics training as well as for planning in Appendix \ref{appdx:model_hyperparameters}.
We outline the hyperparameters for all the module of \ours{} in Table \ref{tab:appdx-hyperparemeters}.

\begin{table}[t]

\centering
\begin{tabular}{lcc}
\toprule
\textbf{Config/Hyperparameter} & \textbf{Value} \\ 
\midrule
\multicolumn{2}{c}{\textbf{Point Completion}} \\
\midrule
jitter prob & 0.3 \\
jitter mean & 0.01 \\
jitter variance & 0.03 \\
global feature channels & 512\\
encoder channels & [32, 64, 128, 256]\\
decoder upsample factors  & [2, 2, 2]\\
decoder hidden channels & 64\\
seed generator blocks & 4\\
seed generator - attn heads & 8\\
seed generator - attn dims & 384\\
seed generator - upsample factor & 2\\
seed generator - output dim & 256\\

\midrule
\multicolumn{2}{c}{\textbf{Point Dynamics model}} \\
\midrule

serialization pattern & Z + TZ + H + TH\\ 
patch interaction & Shift Order + Shuffle Order \\
positional encoding &  [CPE, RPE, CPE, RPE] \\
stride & [2, 2] \\
encoder depth & [2, 2, 2] \\
encoder channels & [32, 128, 256] \\
encoder num heads & [8, 8, 8] \\
encoder patch size &  [256, 48, 256] \\
decoder depth & [2, 2] \\
decoder channels & [128, 256] \\
decoder num heads & [8, 8] \\
decoder patch size & [256, 48] \\
mlp ratio & 4 \\
qkv bias & True \\
drop path & 0.3 \\
pre norm & True \\
shuffle orders & True \\

\midrule
\multicolumn{2}{c}{\textbf{MPPI Planner}} \\
\midrule

planning horizon & 20-30 steps \\
\# num samples & 200 \\
\# top candidates & 64 \\
lambda\_ & 0.7 \\
noise $\sigma$ & 5 \\
\midrule
\multicolumn{2}{c}{\textbf{Training Hyperparameters}} \\
\midrule
loss & HuberLoss, delta=0.25 \\
optimizer & AdamW \\
learning rate & 0.0001 \\
scheduler & StepLR, step\_size=30, $\gamma=0.1$\\
\bottomrule
\end{tabular}%
\caption{Hyperparameter values for different modules in our method.}
\label{tab:appdx-hyperparemeters}

\end{table}

\textbf{SAM3 Prompting}: Given the task description, we first prompt Gemini Robotics 1.5 VLM \cite{team2025gemini} to identify task-relevant objects in the scene by outputting their 2D point locations. We then use these points as prompts to SAM3 \cite{carion2025sam3segmentconcepts} to segment the corresponding objects.

We use the following prompt to instruct the vision-language model to
identify task-relevant objects and provide 2D point prompts for
segmentation:

\begin{vlmprompt}
You are a vision-language model assisting a robot with a manipulation task.

Given:
1. A natural-language task description.
2. An RGB image of the current scene.

Your goal is to identify all objects that are relevant to successfully
completing the task and provide one or more 2D point locations for each
relevant object.

Instructions:

1. Read the task description carefully and determine which physical
objects the robot must interact with, manipulate, move, avoid, or
otherwise reason about to complete the task.

2. In the image, locate each task-relevant object.

3. For each relevant object, select a 2D point that lies clearly INSIDE
the visible region of that object.
   - The point should be near the center of the visible portion of the
     object whenever possible.
   - Do not place the point on the object's boundary.
   - Do not place the point on the background, another object, or an
     occluding object.
   - Prefer points that are visually unambiguous and far from object
     boundaries.
   - If an object is partially occluded, place the point on the visible
     portion of the object.

...

Output format:

[
  {
    "object": "<short object description>",
    "point": [x, y]
  },
  ...
]

Example:

Task: "Pick up the red mug and place it inside the bowl."

Output:
[
  {
    "object": "red mug",
    "point": [312, 241]
  },
  {
    "object": "bowl",
    "point": [487, 326]
  }
]

Be precise: every point must fall inside the corresponding object's
visible pixels in the image.
\end{vlmprompt}

\subsection{Cost Function Specification}
\label{appdx:cost_fn}

One of the key goals of learning a point based world model in the observation space, is to be able to specify cost functions to enable planning. This requires that the rollouts produced by the dynamics model are of high fidelity -- so that the cost/reward computation on the rolled out point clouds is accurate. Since we perform point completion, which enables high fidelity long-horizon rollouts, we can use cost functions which can be defined on point clouds directly. We prompt Gemini 2.5-Flash \cite{comanici2025gemini} with the following prompt to obtain cost functions that can be defined in the point cloud observation space. 

\begin{vlmprompt}
    You are a vision-language model assisting a robot with a manipulation task.

Your goal is to generate a task-specific cost function for evaluating predicted robot states during trajectory planning.

You are given:

1. A natural-language task description:

   <TASK>

2. A JSON file containing objects identified as potentially relevant to the task:
   <OBJECTS_JSON>

The objects JSON contains the objects detected in the scene and their 2D image point locations. It may contain:

* Objects that must be manipulated to complete the task.
* Objects that serve as destinations or targets.
* Objects that provide contextual information needed to complete the task.
* Objects that are not explicitly mentioned in the task but may be relevant because the robot could collide with them or they may constrain the robot's motion.

The generated cost function will be evaluated on 3D point clouds predicted by a robot world model. The point clouds represent the predicted geometry of objects and robot components at future states along candidate trajectories.

### Cost function interface

Generate a Python/PyTorch function with the following interface:

```python
def cost_fn(cfg, point, info):
    """
    Task-specific cost function.

    Args:
        cfg: Configuration object. In particular, cfg.device specifies
             the PyTorch device on which tensors should be created.

        point: Dictionary containing predicted 3D point clouds for
               objects and robot components. Keys correspond to object
               names from the provided objects JSON and may also include
               robot components such as 'left_gripper' and
               'right_gripper'.

               Each point cloud is a PyTorch tensor containing 3D
               coordinates.

        info: Dictionary containing additional task/environment
              information that may be required to evaluate the task,
              such as goal positions, grasp/contact state, or task
              stage. Additional information is available under
              info['extra'] when provided.

    Returns:
        A single scalar PyTorch cost. Lower cost indicates a more
        desirable state.
    """
    ...
    return total_cost
```

### Requirements

1. **Infer the task objective**

   Carefully interpret `<TASK>` and determine what the robot must accomplish, including any intermediate stages or object relationships required for successful completion.

2. **Use the objects JSON**

   Use the provided objects to determine which objects should influence the cost.

   Do not invent task objects that are not present in the JSON.

3. **Include potentially relevant objects**

   Do NOT include clearly irrelevant scene elements such as the background or image artifacts. However, include objects that could potentially be relevant to execution, even if they are not explicitly mentioned in the task.

   In particular, consider objects that the robot arm, gripper, or manipulated object could collide with or that could constrain the robot's motion.

4. **Operate on predicted point clouds**

   The cost function should primarily operate on the 3D point clouds in `point`.

   For example, object geometry may be summarized using:

   * Point-cloud centroids.
   * Distances between point clouds.
   * Distances between object centers.
   * End-effector-to-object distances.
   * Object-to-target distances.
   * Relative positions.
   * Spatial overlap or containment.
   * Proximity to potential obstacles.

   Choose geometric quantities appropriate for the specific task.

5. **Use `info` when necessary**

   Use `info` for information that cannot be obtained reliably from the point clouds, such as:

   * Whether an object is currently grasped.
   * Task stage.
   * Goal positions.
   * Other task-specific environment state.

   Do not assume privileged information is available unless it is provided through `point` or `info`.

6. **Measure task progress**

   The cost should provide a meaningful measure of progress toward task completion.

   For multi-stage tasks, the cost may change depending on the current task stage. For example, it may first encourage reaching an object and then encourage moving the object toward its destination after grasping.

7. **Collision avoidance**

   Include collision or near-collision costs when potentially relevant objects could interfere with the robot's execution.

   The collision cost should increase as the robot or manipulated object approaches a potential obstacle.

   Collision avoidance should complement, rather than overwhelm, the primary task objective.

8. **Smoothness**

   Prefer smooth cost functions that provide useful ranking information between candidate states, including when none of the candidates completely succeeds.

   Prefer continuous functions such as Euclidean distances, `torch.tanh`, exponentials, or other smooth distance-based penalties over hard binary thresholds when appropriate.

9. **Scalar output**

   The function must return a single scalar PyTorch tensor:

   ```python
   return total_cost
   ```

   Lower cost must correspond to a better state.

10. **Task-specific design**

    Do not produce a generic cost function. Construct the cost specifically from the provided task description and objects JSON.

11. **Keep it simple**

    Use only the terms necessary to capture the essential task objective and relevant collision constraints. Avoid unnecessarily complicated heuristics.

12. **PyTorch compatibility**

    Use PyTorch operations for geometric calculations so that the function can operate directly on the tensors in `point`.

### Output format

Return:

1. A brief description of the task objective and the relevant object roles.

2. A brief description of the cost terms and what each term measures.

3. The complete Python/PyTorch implementation:

```python
def cost_fn(cfg, point, info):
    ...
    return total_cost
```

4. A brief list of assumptions about the available keys in `point` and `info`.

The final implementation must be directly usable as a task-specific cost function for ranking predicted robot trajectories.

\end{vlmprompt}

Alternatively, manually constructed cost functions can also be used. We show an example of the cost function adapted in a very straightforward manner from ManiSkill \cite{Tao2024ManiSkill3GP} in a pytorch-like code \ref{lst:pickcube_cost}.

\begin{lstlisting}[style=pythonstyle, caption={Cost function for the PickCube task that operates on \ours{}'s rolled out point clouds.}, label={lst:pickcube_cost}]
def cost_fn_pick_cube(cfg, point, info):
    """
    Cost function for the PickCube-v1 task.
    """

    goal_xyz = torch.tensor(info['extra']['goal_pos'], device=cfg.device)

    left_gripper_xyz = point['left_gripper']
    right_gripper_xyz = point['right_gripper']
    cube_xyz = point['cube']

    cube_center_xyz = cube_xyz.mean(dim=0)
    tcp_xyz = (left_gripper_xyz.mean(dim=0) +
               right_gripper_xyz.mean(dim=0)) / 2

    tcp_to_cube_dist = torch.norm(cube_center_xyz - tcp_xyz, dim=0)
    reaching_reward = 1.0 - torch.tanh(5.0 * tcp_to_cube_dist)

    cube_to_goal_dist = torch.norm(cube_center_xyz - goal_xyz, dim=0)
    place_reward = 1.0 - torch.tanh(5.0 * cube_to_goal_dist)

    contact_reward = 1.0 if info['grasped'] else 0.0

    if info['grasped']:
        total_cost = -place_reward
    else:
        total_cost = -(reaching_reward + contact_reward)

    return total_cost
\end{lstlisting}

Since the notion of contacts is not baked in explicitly, we perform grasping when the gripper is within $2.5$ cm of the object (i.e distance between \texttt{tcp\_xyz} and \texttt{cube\_center\_xyz} is $\leq 0.025$) and assume a success (i.e \texttt{info['grasped'] = True}). Alternatively, we can also try computing the intersection between the left-gripper, object and the right-gripper to determine a contact, however, in practice we found it hard to tune the threshold for detecting the contact. We leave addressing this issue in point world models as a direction for future work. 

\subsection{Failure Scenarios}
\label{appdx:failure_cases}
\begin{figure}[!t]
\centering
\includegraphics[width=0.9\linewidth]{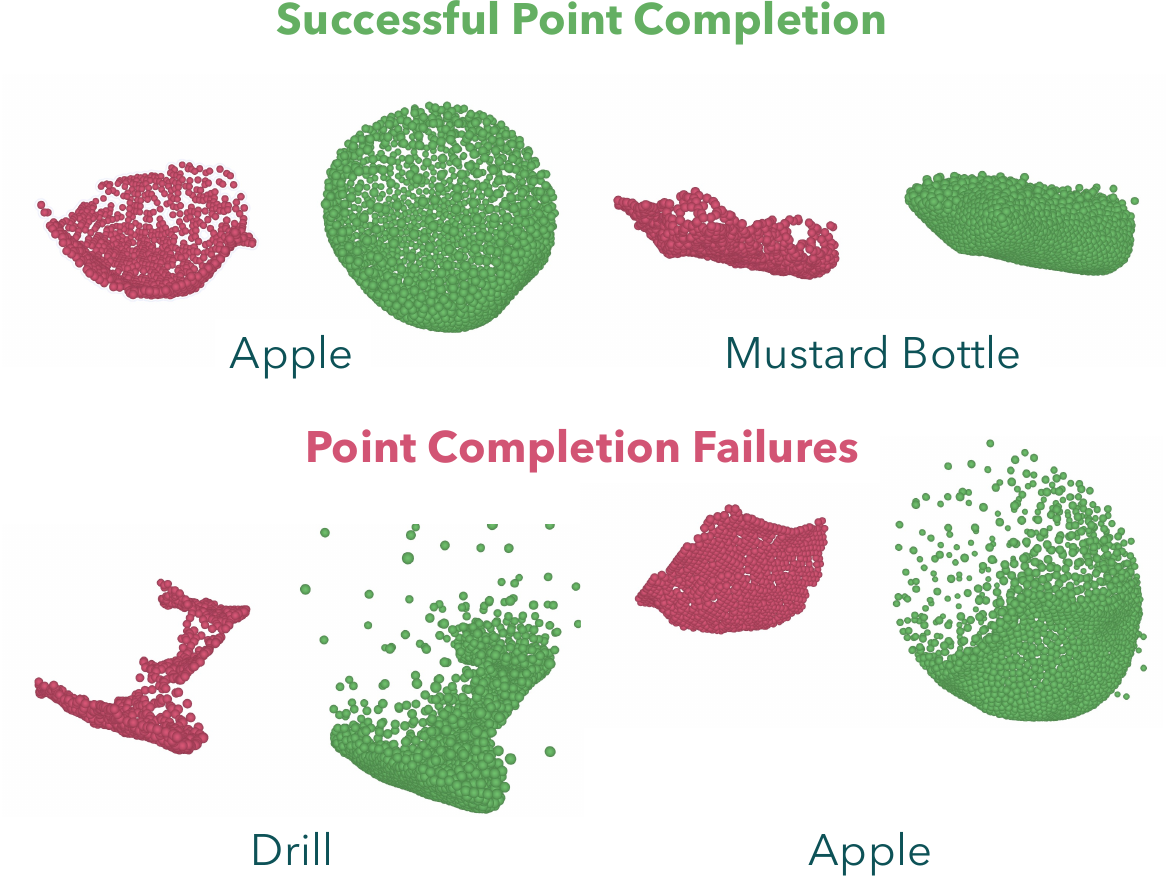}
\caption{\textit{Point completion success \& failures}: We show few samples of the fine-tuned point cloud completion model. \textbf{(Top)} Successful point cloud completions on partial point clouds of apple and mustard bottle. \textbf{}{(Bottom)} Failure scenarios of the completion on Drill and Apple from the YCB dataset.}
\label{fig:point_completion_failures}
\end{figure}

\textbf{Point completion.} First we discuss some of the failures of point cloud completion module on real world objects. We show a sample of completion success as well as failures in Figure \ref{fig:point_completion_failures}. The failure scenarios primarily arise due to a mismatch between the training partial point cloud distribution and the real-world partial point cloud distribution. As mentioned in Sec.~\ref{sec:limitations}, our current data generation pipeline for point cloud completion does not include occlusions cause by other objects in the scene. This is because we generate partial point clouds of the objects by considering individual meshes of the objects and generating depth maps from 100 random viewpoints and backprojecting the depth to obtain the point cloud. This process, by design, does not include various kinds of occlusions caused due to the presence of other objects.
Further, the noisy generation of points outside the boundary as seen in Fig \ref{fig:point_completion_failures} is also a common failure point in the pipeline.

\subsection{Additional Experiments on \ours{} robustness towards lighting conditions and camera viewpoints}
\label{appdx:lighting_camera}

\begin{figure}[!t]
\centering
\includegraphics[width=0.9\linewidth]{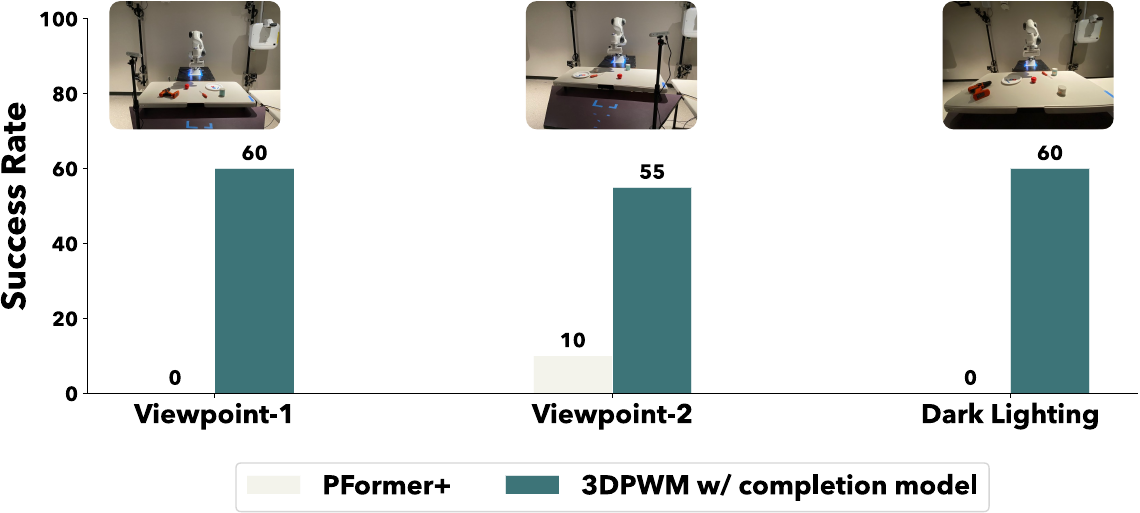}
\caption{\textit{Lighting \& Viewpoint condition}: We tested the robustness of \ours{} on \texttt{PickApple} task with two additional viewpoints as well as darker lighting condition. (Each plot is averaged over 20 trials.)}
\label{fig:lighting_viewpoint}
\end{figure}

We also experimented with 2 additional camera viewpoints as well as changing the lighting conditions to test the robustness of \ours{}. It has been shown that policies trained with calibrated point clouds are generally more robust to such lighting and camera viewpoint changes \cite{zhu2022viola, PCWM}. This was further corroborated by our results on \texttt{PickApple} shown in Figure \ref{fig:lighting_viewpoint}. Specifically, we observe that \ours{} maintains its higher performance when compared with \pformerplus.

\subsection{Additional Experiments: Comparison to GNN based methods}
\begin{table}[!h]
\begin{center}
\caption{Comparison between 3DPWM and graph neural network GBND}
\label{tab:gbnd}
\centering
\renewcommand*{\arraystretch}{0.35}
\setlength{\tabcolsep}{2pt}
\resizebox{0.65\linewidth}{!}{
\begin{tabular}{l >{\centering\arraybackslash}p{5em}>{\centering\arraybackslash}p{3.5em} | >{\centering\arraybackslash}p{5em}>{\centering\arraybackslash}p{3.5em}}
\toprule
\multirow{2}{*}[-0.5em]{\textbf{Task}} 
& \multicolumn{2}{c}{\textbf{ \scriptsize Chamfer L1 ($\downarrow$)}} 
& \multicolumn{2}{c}{\textbf{ \scriptsize Reward Corr $(r)$ ($\uparrow$)}} \\
\cmidrule(lr){2-3} \cmidrule(lr){4-5}
 & GBND \cite{zhang2024dynamic} & \ours{} & GBND \cite{zhang2024dynamic} & \ours{} \\
\midrule
 PickCube & 0.185 & \cellcolor{best}0.009 & 0.37 & \cellcolor{best}0.86\\
 Pick YCB & 0.21 & \cellcolor{best}0.028 & 0.33 & \cellcolor{best}0.85 \\
 Box Cleanup & 0.301 & \cellcolor{best}0.022 & 0.16 & \cellcolor{best}0.83 \\
\bottomrule
\end{tabular}}
\end{center}
\end{table}

\noindent Gaussian Splatting based methods like \cite{zhang2024dynamic} 
comprise of (a) Gaussian point-based dynamics and 
(b) splatting-based rendering. Since \ours{} focuses on modeling
the dynamics and \emph{not} the rendering, we compare against 
the GNN-based dynamics model from \cite{zhang2024dynamic} in 
Table \ref{tab:gbnd}. As shown in \cite{huang2025particleformer} 
and confirmed by our results -- GNN-based dynamics 
underperforms transformer-based approaches. 
Integrating a rendering pipeline into \ours{} remains a future work 
that we wish to address with \ours{} as the dynamics model.

\subsection{Additional Experiments: Sim2Real in Cluttered Environment}
\begin{figure}[!h]
\centering
\includegraphics[width=0.9\linewidth]{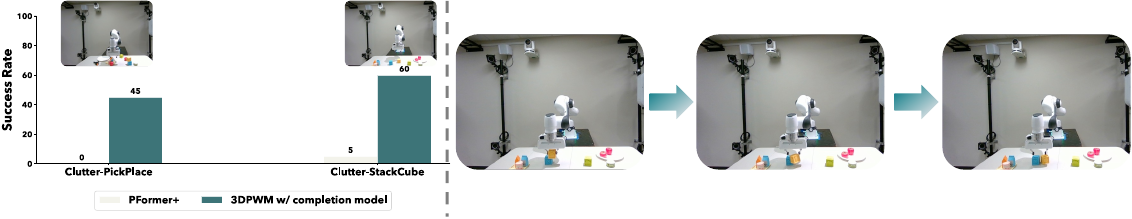}
\caption{\textit{Sim2Real on Cluttered Environments}: \textbf{(Left)} We evaluate \ours{} on two cluttered-environment tasks:
    \texttt{PickPlaceApple}, in which the robot must pick an apple
    surrounded by clutter, and \texttt{StackCube}, in which the robot
    must stack cubes amid a cluttered scene.
    (Results are averaged over 20 trials per task.) \textbf{(Right)} Failure scenarios of \ours{} on \texttt{StackCube} occurs at the very end during placement of the yellow cube on top of the blue cube.}
\label{fig:clutter}
\end{figure}

We further evaluate the planning capabilities of \ours{} in cluttered environments across two tasks: \texttt{PickPlaceApple} and \texttt{StackCube}. In \texttt{PickPlaceApple}, the apple is initialized near a group of cluttered objects, and the cost function incorporates a heuristic penalty for collisions with surrounding objects. \texttt{StackCube} requires stacking a yellow cube on top of a blue cube in a moderately cluttered environment. This task is particularly challenging, as precise placement is critical; notably, all observed failures on \texttt{StackCube} are near-misses, as illustrated in Figure \ref{fig:clutter} (right).

\subsection{Baselines: Particleformer \cite{huang2025particleformer}}
\label{appdx:baselines}

Here we lay out the specific details of our re-implementation of ParticleFormer \cite{huang2025particleformer}. Each partial point input has point velocities and one-hot vector representing instance segmentation.  We run two architecture variants (i) ParticleFormer$^*$, which is re-implementation of the the ParticleFormer baseline with 3 transformer layers as in the original paper, and (ii) \pformerplus an equivalent variant of encoder and decoder that uses our PTV3 architecture \cite{ptv3}. Unlike \ours{}, we train both these variants on segmented partial point clouds and supervise the dynamics model with the hybrid loss (a combination of Chamfer and Hausdorff losses) as proposed by the authors.

We report the performance of ParticleFormer$^*$ here in Table \ref{tab:pformer_additional_expt} along with the copied results of \pformerplus from Table \ref{tab:rollout_metrics}. We find that the original ParticleFormer with 3-layers of transformer is insufficient to capture the dynamics of both \texttt{PickCube} as well as \texttt{Box Cleanup} tasks and fares lower to \pformerplus.

\begin{table}[!t]
\centering
\caption{
Chamfer distance and reward correlation on both variants of ParticleFormer ($+$ and $*$)
(\texorpdfstring{\textcolor{best}{Best} and \textcolor{secondbest}{second best}}{Best and second best methods are highlighted}).
}
\label{tab:pformer_additional_expt}
\renewcommand*{\arraystretch}{1.15}
\setlength{\tabcolsep}{3pt}
\resizebox{\linewidth}{!}{
\begin{tabular}{l l c c c | c c c}
\toprule
\multirow{2}{*}{\textbf{Task}} 
& \multirow{2}{*}{\textbf{\scriptsize\makecell{Avg. \\ rollout len.}}}
& \multicolumn{3}{c}{\textbf{Chamfer L1 ($\downarrow$)}} 
& \multicolumn{3}{c}{\textbf{Reward Corr $(r)$ ($\uparrow$)}} \\
\cmidrule(lr){3-5} \cmidrule(lr){6-8}
& & PFormer$^+$ & \makecell{PFormer$^*$} & \ours{} 
  & PFormer$^+$ & \makecell{PFormer$^*$} & \ours{} \\
\midrule
PickCube 
& 66 
& \cellcolor{secondbest}0.080 
& 0.132 
& \cellcolor{best}0.009
& \cellcolor{secondbest}0.61 
& 0.41 
& \cellcolor{best}0.86 \\
Box Cleanup 
& 218 
& \cellcolor{secondbest}0.132 
& 0.225 
& \cellcolor{best}0.022 
& \cellcolor{secondbest}0.30 
& 0.19 
& \cellcolor{best}0.83 \\
\bottomrule
\end{tabular}}
\end{table}

\label{appdx:environments}
\subsection{Environments} 
\begin{figure}[!h]
\centering
\includegraphics[width=0.75\linewidth]{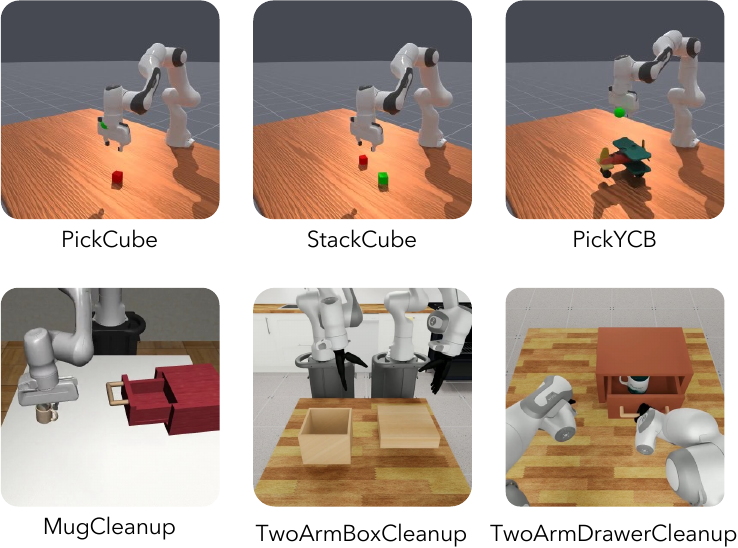}
\caption{\textit{Environments used to evaluate \ours{}}:}
\label{fig:environments}
\end{figure}

We evaluate our dynamics model design across 6 different tasks -- \texttt{PickCube}, \texttt{StackCube} and \texttt{PickYCB} from the Maniskill simulator \cite{Tao2024ManiSkill3GP}, \texttt{MugCleanup} from MimicGen \cite{mandlekar2023mimicgen} and \texttt{TwoArmBoxCleanup} and \texttt{TwoArmDrawerCleanup} from Dexmimicgen \cite{jiang2025dexmimicgen}. A sample snapshot of the environment is shown in Figure \ref{fig:environments}.
\label{appdx:environments}

\xhdr{\textbf{PickCube}:} Lift the cube from the table to the goal position in the air specified by a 3D point in space. \newline
\textbf{StackCube}: Stack the red cube on top of green cube. \newline
\textbf{PickYCB}: Pick up the YCB object from the table to the goal position in the air specified by a 3D point in space.  \newline
\textbf{MugCleanup}: Pick and place the mug in side the drawer and close it. \newline
\textbf{TwoArmBoxCleanup}: Pick up the lid of the box using two dexterous hands and place it on the box. \newline
\textbf{TwoArmDrawerCleanup}: Pick and place the mug using two dexterous hands in side the drawer and close it.

For all the tasks with parallel jaw gripper, we use the end-effector delta actions, and for dexterous tasks we use 24-dim end effector actions (6-dim for pose, and 6 for finger control on each arm).

\begin{table}[!h]
\centering

\centering
\begin{tabular}{lccc}
\toprule
\textbf{Task} & \textbf{Observation Points} & \textbf{Action dim} \\ \midrule
\texttt{PickCube} & $1200$ & $7$ \\
\texttt{StackCube} & $1500$ & $7$ \\
\texttt{PickYCB} & $1200$ & $7$ \\ 
\texttt{MugCleanUp}& $8192$ & $7$ \\ 
\texttt{TwoArmBoxCleanup} & $10000$ & $24$ \\ 
\texttt{TwoArmDrawerCleanup} & $12000$ & $24$ \\ 
\midrule
\end{tabular}%
\caption{Number of points for complete point cloud and action spaces of each of the tasks.}
\label{tab:appdx-state-action-space}
\end{table}
\label{appdx:add_expts_tried}
\subsection{Additional experiments we tried that did not work.}
In this section we lay out some of the additional experiments we 
tried but did not see success and believe that this would be helpful
to share with the community to improve upon.

\textbf{Using SAM3D instead of custom trained point completion.} We 
experimented with SAM3D \cite{chen2025sam3d} but found that the completions,
while generalizable to various commonly found objects \emph{do not align}
with the partial input despite providing sensor depth from the camera (
instead of using DepthAnything or other foundation depth models).
This is a known issue (see Issues \href{https://github.com/facebookresearch/sam-3d-objects/pull/79#issuecomment-3597044752}{$\#79$} and \href{https://github.com/facebookresearch/sam-3d-objects/issues/162}{$\#162$} on the SAM3D github
repository.)

\textbf{Using Depth Foundation models to avoid noisy completions.} We 
also tested using MOGE-2 \cite{wang2026moge} and DepthAnything3 \cite{lin2025depth} since they have shown
impressive performance for depth estimation. However, we observed that
while the depth was very clean and so were the
resulting backprojected point clouds -- these models still suffered 
with two major challenges which made decision making difficult: (a) similar
to SAM3D, these depth foundation models suffered with object misalignment
in the real world, and (b) they tended to miss out smaller details (depending
on the viewpoint of the object) of the 
objects that could potentially be crucial for manipulation.

\end{document}